\documentclass[conference]{IEEEtran}
\IEEEoverridecommandlockouts
\usepackage{cite}
\usepackage{amsmath,amssymb,amsfonts}
\usepackage{algorithmic}
\usepackage{graphicx}
\ifCLASSOPTIONcompsoc
    \usepackage[caption=false, font=normalsize, labelfont=sf, textfont=sf]{subfig}
\else
\usepackage[caption=false, font=footnotesize]{subfig}
\fi
\usepackage{textcomp}
\usepackage{xcolor}
\def\BibTeX{{\rm B\kern-.05em{\sc i\kern-.025em b}\kern-.08em
    T\kern-.1667em\lower.7ex\hbox{E}\kern-.125emX}}
\begin{document}

\title{Automatic counting of mounds on UAV images: combining instance segmentation and patch-level correction\\
%{\footnotesize \textsuperscript{*}Note: Sub-titles are not captured %in Xplore and
}

\author{\IEEEauthorblockN{1\textsuperscript{st} Majid Nikougoftar Nategh}
\IEEEauthorblockA{\textit{Concordia Institute for Information Systems Engineering} \\
\textit{Concordia University}\\
Montreal, Canada \\
nikougoftar.majid@gmail.com}
\and
\IEEEauthorblockN{2\textsuperscript{nd} Ahmed Zgaren}
\IEEEauthorblockA{\textit{Concordia Institute for Information Systems Engineering} \\
\textit{Concordia University}\\
Montreal, Canada \\
ahmed.zgaren84@gmail.com}
\and
\IEEEauthorblockN{3\textsuperscript{rd} Wassim Bouachir}
\IEEEauthorblockA{\textit{\phantom{.................--.....}Department of Science and Technology \phantom{.................}} \\
\textit{Université TELUQ}\\
Montreal, Canada \\
wassim.bouachir@teluq.ca}
\and
\IEEEauthorblockN{4\textsuperscript{th} Nizar Bouguila}
\IEEEauthorblockA{\textit{Concordia Institute for Information Systems Engineering}\\
\textit{Concordia University}\\
Montreal, Canada \\
nizar.bouguila@concordia.ca}
%\and
%\and
}

\maketitle

\begin{abstract}
Site preparation by mounding is a commonly used silvicultural treatment that improves tree growth conditions by mechanically creating planting microsites called mounds. Following site preparation, the next critical step is to count the number of mounds, which provides forest managers with a precise estimate of the number of seedlings required for a given plantation block. Counting the number of mounds is generally conducted through manual field surveys by forestry workers, which is costly and prone to errors, especially for large areas. To address this issue, we present a novel framework exploiting advances in Unmanned Aerial Vehicle (UAV) imaging and computer vision to accurately estimate the number of mounds on a planting block.
The proposed framework comprises two main components. First, we exploit a visual recognition method based on a deep learning algorithm for multiple object detection by pixel-based segmentation. This enables a preliminary count of visible mounds, as well as other frequently seen objects (e.g. trees, debris, accumulation of water), to be used to characterize the planting block. Second, since visual recognition could limited by several perturbation factors (e.g. mound erosion, occlusion), we employ a machine learning estimation function that predicts the final number of mounds based on the local block properties extracted in the first stage. We evaluate the proposed framework on a new UAV dataset representing numerous planting blocks with varying features. The proposed method outperformed manual counting methods in terms of relative counting precision, indicating that it has the potential to be advantageous and efficient in difficult situations.

\end{abstract}

\begin{IEEEkeywords}
Object counting, Mound detection, UAV imagery, Instance segmentation, Mask-RCNN, Computer vision, Precision forestry
\end{IEEEkeywords}

\section{Introduction}
Mechanical site preparation by mounding has been promoted and recognized as a popular technique in forest industry due to the abiotic and biotic characteristics of North American terrains. Mounding is also referred to a silvicultural technique for constructing elevated planting spots that are free of water logging, and with little vegetation competition in the soil \cite{b1}. A key issue after mounding is to precisely estimate the number of mounds created on each planting block, which corresponds to the number of tree seedlings to be planted.

Manual counting is a commonly used method in forest industry. To do so, forest workers count mechanically prepared mounds on a section of the site and use the result to estimate a total number for the entire site, assuming that mound density remains constant on a given plantation block. However, this approach is time consuming, expensive, and prone to human error. Furthermore, mound density often varies on the same block depending on the characteristics of each zone. Motivated by recent advances in sensor technology used in drone platforms for data collection, forestry managers also used visual interpretation of Unmanned Aerial Vehicle (UAV) images as an alternative to field manual counting. Image interpretation and analysis are thus performed by human operators, in order to detect, identify, and count mounds on UAV orthomosaics. However, this requires a skilled human interpreter, and due to perception variation among humans on the nature of the objects, data quality, and scale from one site to the next. 

%Potentially, counting mounds has traditionally required human involvement, either through field work including grid searching of microsites or image interpretation of aerial imagery. Furthermore, in order to minimize human error and expedite the planting of seedlings, expert observers with a significant time commitment for hard labour are required, which make these approaches less effective. Moreover, the size and diversity of the terrains is another challenge that places further %constraints.

The purpose of our work is to develop a computer vision framework based on a UAV platform to make human work easier and more efficient. We propose a new method for mound counting a combination of two models: 1) local image segmentation using deep learning methods and 2) patch-level correction by applying regression models. First, the local segmentation model is trained on UAV images that have been manually annotated to detect and segment mounds, as well as other relevant objects, including trees, woody debris and water accumulations. In fact, pixel-wise instance segmentation is used to count visual mounds and other objects of interest using local image segmentation as a first stage. The first processing step is essential in our framework, in order to characterize each image region by quantifying the presence of relevant object instances. Second, a patch-level correction model is applied to produce the final prediction of mound count, based on preliminary object count. Since mounds can be destroyed or occluded following their creation (e.g. due to erosion, presence of debris and trees), we cannot rely only on visual detection to estimate their number). Therefore, we formulate the task of counting mounds using a sequential approach that includes local pixel-wise object segmentation and patch-level correction.

We evaluated our framework on a dataset of UAV orthmosaics with different properties. The obtained results emphasize the importance of using both models sequentially.

The rest of this article is organized as follows. Section II introduces background concepts and related works. Section III, provides a detailed description of the proposed framework. Experimental results are presented in section IV. Finally, section V concludes the paper.

\section{Related works}
Visual object counting, also known as crowd counting, is a computer vision task that encompasses all issues and challenges associated with estimating the number of times a specific object appears in an image \cite{b2}. Methods towards object counting in the literature can be divided into three categories based on the features used: traditional approaches, deep learning approaches, and hybrid approaches.

\subsection{Traditional crowd counting approaches}
Traditional crowd counting methods rely on hand-crafted features and are categorised into two categories based on the extracted feature category: direct detection methods and indirect detection methods.

Direct detection method (also known as detection-based method) has been widely used in many approaches\cite{b3, b4}, \cite{b6, b7}. These approaches localize the position of each object in a single input image and the number of detections is subsequently used as the crowd count. Traditionally, low-level features including Haar wavelets \cite{b4}, histograms of oriented gradients (HOG) \cite{b3}, edgelet\cite{b6}, and shapelet \cite{b8} are used as region descriptors. Then a mainstream classifiers such as Support Vector Machine (SVM) \cite{b9}, boosted trees \cite{b10} and random forests \cite{b11} is trained for classification. Finally, the number of object instances that the classifier produces on a test image is considered as the crowd count. Although detection-based methods have been effectively employed in low-density crowds, their performance decreases substantially when applied in high-density crowds with small and obscured objects, since they are based on low-level features.

Unlike direct techniques, object counting with indirect approach (also known as feature-based method or regression method) starts with taking the entire crowd as an object, extracting local, global and texture features of the crowd, and then establishing a mapping to the number of dense crowds to estimate the number of crowds indirectly \cite{b5}. These approaches have the advantage of not relying on the learning detector\cite{b5}, and they are more effective because recognition is based on features rather than objects. However, regression-based counting directly maps from the extracted features of images to the number of objects, and because they ignore the object distribution information within the region, these methods do not have the ability to explicitly detect and localize each object \cite{b12}. 
Although density map regression-based approaches have achieved significant progress, they are still inadequate for real-world applications, particularly when large-scale variation is present. In addition, they are not capable to capture semantic information since they rely on low-level features. %Furthermore, because these methods do not incorporate spatial information, accurate location and size of objects in a scene cannot be predicted.

Traditional crowd counting methods are generally successful in moderately crowded scenes and are faster to process because they do not require considerable computational resources. However, they require a large number of training sets in order to have an effective training model. The applicability of these methods is also limited, and they are ineffective in dense crowds scenes with severe occlusion.

\subsection{Deep learning approaches}
Deep learning models have outperformed traditional machine learning approaches in recent years in the field of crowd counting. To learn and classify crowd regions of an image, deep learning algorithms rely on deep neural networks to extract semantic invariant features. Therefore, current research has shifted its focus to developing CNN-based techniques, as CNNs provide a more robust feature representation, compared to the hand-crafted features utilized in traditional approaches.

%Basic CNN approaches integrate basic CNN layers to proposed first deep learning models for crowd counting. Then, advanced architectures have been incorporated to mitigate scale variation errors which are based on multi-column or pyramid architectures.

\subsubsection{Basic CNN approaches}
These methods incorporate basic CNN layers as initial deep learning approaches for crowd counting. For counting people in highly dense crowds, Wang et al. \cite{b16} developed an end-to-end deep CNN regression model. In their architecture, they modified the original AlexNet network \cite{b17} by replacing the final fully connected layer with a single neuron to obtain an object count. Furthermore, training data augmented with additional negative samples are used to eliminate false responses in the backdrop of the images. Fu et al. \cite{b18} presented an optimized CNN approach based on the multi-stage ConvNet to estimate crowd density. Their optimization method is centered on removing some network connections based on the similar feature maps. Crowd images were then classified into one of five classes using two CNN cascade classifiers: very high density, high density, medium density, low density, and very low density. %Inspired by their success, Walach et al. \cite{b20} used selective sampling methods and layered boosting, which involves iteratively adding CNN layers to the model with the purpose of assessing the residual error of the initial prediction.

\subsubsection{Scale-aware CNN approaches}
Single scale CNN models are less effective at accurately predicting density maps due to image scale variance. As a result of the foregoing constraint, basic CNN-based techniques evolved into more sophisticated models that were scale-resistant. To capture both high-level semantic information and low-level features, Boominathan et al. \cite{b20} combined deep and shallow fully convolutional networks to address crowd scales as well as perspective variations. %In \cite{b22}, a scale-aware counting model (HydraCNN) developed to estimate object densities in a variety of crowded scenarios without any explicit geometric information of the scene, and a multiscale nonlinear regression model used to predict crowd density using a pyramid of image patches extracted at multiple scales.
With the adoption of CNN-based density map regression methods, more advanced CNN architectures by incorporating multi-column networks
such as Multi-column CNN architecture (MCNN) \cite{b22}, Switched CNN (Switch-CNN) \cite{b23}, and Congested Scene Recognition Network (CSRNet) \cite{b24} have seen a remarkable improvement in performance. 
%Multi-column CNN architecture (MCNN) \cite{b22} uses multi-column deep neural networks with different convolution kernel sizes to learn different scales of features. Switched CNN (Switch-CNN) \cite{b23} performs a fusion of three single CNNs and a selector to transmit crowd scene patch to the best independent CNN regressor. Congested Scene Recognition Network (CSRNet) \cite{b24} based on dilated kernels has been used to generate high quality density maps and capture large scale information instead of using a multi-branch architecture.

\subsubsection{Context-aware CNN approaches}
Context-aware models were designed with the aim of reducing estimation errors by combining local and global contextual information into the CNN architecture. By using different coefficient weights, Sheng et al. \cite{b25} proposed a generalized variant of weighted-VLAD. To do so, semantic information was incorporated into learning locality-aware feature (LAF) sets designed to investigate the spatial context and local information of crowds. In \cite{b26}, a count estimation method based on an end-to-end CNN architecture was developed. Instead of partitioning the image into patches, the final crowd count outputs the entire image in this method. As a result, due to the shared computations on overlapping regions achieved by integrating multiple stages of processing, the complexity is reduced.

\subsection{Hybrid approaches}
Hand-crafted features and deep features are the two most common feature representations used in hybrid approaches. Lin et al. \cite{b27} proposed a low-cost method for counting people in videos. To train a small Local Binary Patterns (LBP) cascade classifier, the suggested architecture leverages a knowledge distillation strategy to transfer knowledge from a CNN object detector. Following the detection of people in video frames using YOLO \cite{b28}, images of people with a confidence level greater than 30\% are used to train the LBP cascade classifier, which is utilised to detect and track pedestrians. In \cite{b29} a combination of two stage detector was proposed to automate the task of detecting and counting the number of planting microsites using multispectral UAV imagery. Object proposals are firstly generated by applying a cascade detector based on LBP features. Then, candidate objects in the second stage were classified by a trained CNN network. Due to the fusion of various features and the use of more than one classifier, hybrid approaches in general may have an excessive computational cost.

\section{Proposed method}
\subsection{Motivations and overview}
In this work, we aim to precisely count the number of mounds on each planting block represented by an orthomosaic. For each planting block, batch of images captured using UAV is reconstructed to produce a high resolution orthomosaic. We divided each orthomosaic into fixed cell sizes due to the high resolution of images, and used them as the input for our framework, as shown in Fig. 1. Visual inspection of different blocks (see Fig. 2) shows that the number of mounds is varying from one patch to another. In fact, this variation is due to many factors, such us mechanical site specificities, environmental factors (dry, wet, and snow), and presence of other objects, such as debris and trees (e.g. mound occlusion by debris, and appearance change due to tree shadows).
%[scale=0.3]
\begin{figure}[htbp]
\centerline{\includegraphics[width=7.1cm,height=3.5cm]{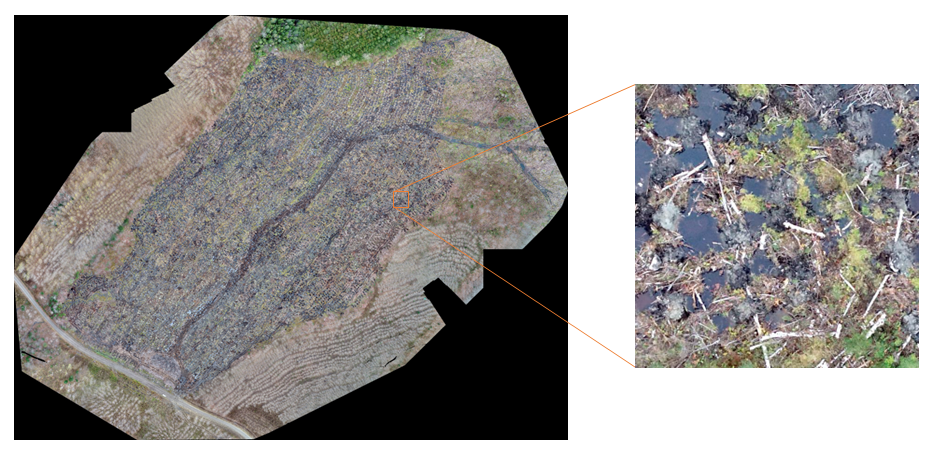}}
\caption{Orthomosaic of one planting block captured and reconstructed  (left) along with the example of one extracted patch (right).}
\label{fig}
\end{figure}

\begin{figure}[htbp]
\centerline{
    \subfloat[]{\label{figur:2}\includegraphics[scale=0.2569]
    {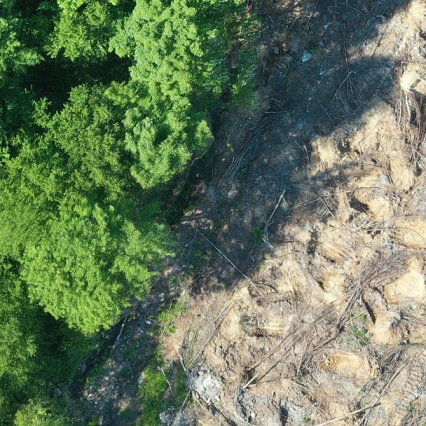}}
    \
    \subfloat[]{\label{figur:2}\includegraphics[scale=0.18]
    {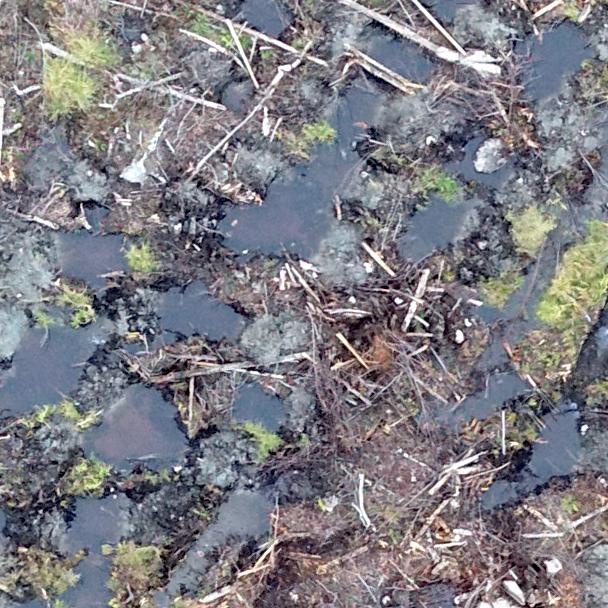}}
     \
    \subfloat[]{\label{figur:2}\includegraphics[scale=0.18]
    {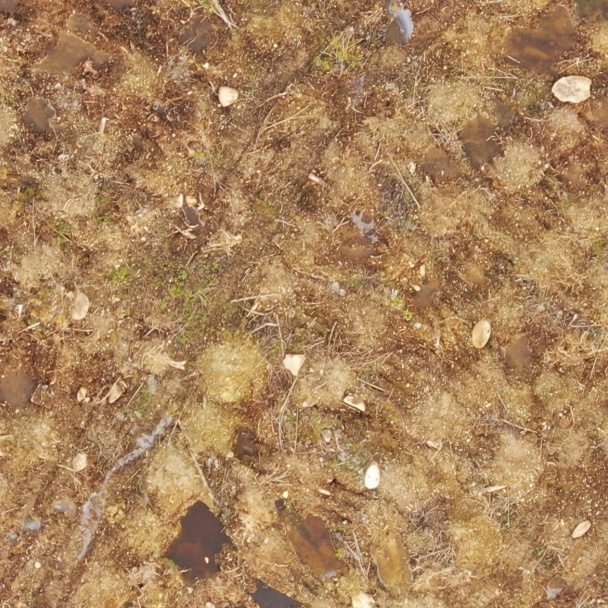}}
    }
\caption{Examples of challenges for three patches from different orthomosaics. (a) Presence of tree shadow causing partial occlusion of mounds. (b) Water accumulation due to heavy rain. (c) Mounds with similar texture to the surrounding areas (background) in dry terrain.}
\label{fig} 
\end{figure}

To handle such difficult factors, a sequential two-step paradigm is used for system training. Firstly, we propose to use an instance segmentation model to detect mounds and to quantify the presence of different objects in local patches. Secondly, we employ patch-level correction to obtain a final number of mounds. The system training procedure is illustrated in Fig. 3.

Once the entire system has been trained, the two models are used to perform mound counting on a new orthomosaic. That is, the local image segmentation is applied as a preliminary method to segment the objects in pixel-level on each patch of a planting block. The results of this step, which include the number of visually detected mounds and the ratio of other objects (e.g, tree, debris and water), are then fed as input features to the second model for a final patch-level correction.

As stated above and explained further, due to the presence of multiple objects and the limitations of occluded mounds, the efficacy of merely performing local object detection method degrades. Therefore, our two-stages strategy is important for achieving accurate and precise counting under our application constraints. The procedure of analyzing a new patch of an orthomosaic to predict mound counting is depicted in Fig. 4. The details of each phase of our framework are presented in the following sub-sections.

\begin{figure}[htbp]
\centerline{\includegraphics[scale=0.39]{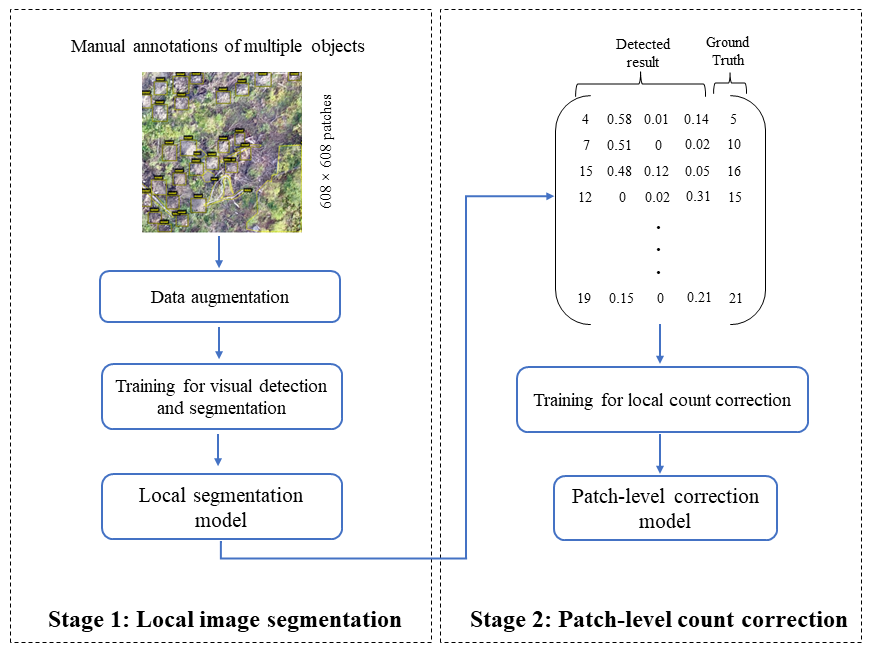}}
\caption{Pipeline of the system training for two stages.}
\label{fig}
\end{figure}

\begin{figure}[htbp]
\centerline{\includegraphics[scale=0.32]{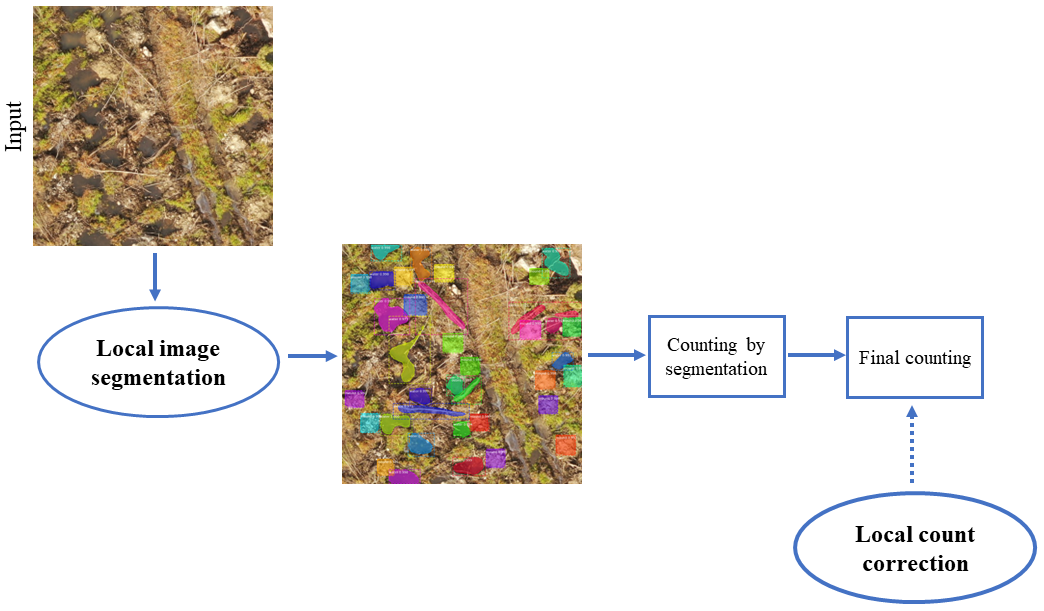}}
\caption{The procedure for evaluating a new patch using our framework.}
\label{fig}
\end{figure}

%In this section we go over the proposed approach for mound counting. Fig. 3 shows different stages of our framework.

%\begin{figure}[htbp]
%\centerline{\includegraphics[width=9cm, height=1.5cm]{Figure3.png}}
%\caption{Pipeline of the proposed framework.}
%\label{fig}
%\end{figure}

\subsection{Local image segmentation}
The first step of our method is to perform local instance segmentation to identify and segment multiple objects in each patch. The main motivation for this step is that different plantation blocks have different properties, and mound density highly depends on block characteristics. In this regard, quantifying the presence of mounds and other objects is an accurate objective indicator to obtain the properties of a bock. In order to accomplish this, local object instance segmentation is used to detect distinct objects belonging to the same category and to assign a unique instance label to the associated pixels.

Instance object segmentation combines object detection, which outputs bounding box coordinates, and semantic segmentation, which outputs segmentation masks. In this work, we use Mask R-CNN \cite{b30} for instance object segmentation to detect mounds and segment all objects in the image. Mask R-CNN is a cutting-edge instance segmentation technique that adds a segmentation mask generating branch to its predecessor, Faster-RCNN, to accomplish proper object detection and pixel-level instance segmentation. We employ Mask R-CNN because it is a two-stage object detector that takes advantage of anchor boxes. Anchor boxes enable this method to detect multiple objects in different scales, which improves the efficiency of the method and provide more accurate localization and classification. Mask R-CNN comprises two stages to produce a final mask segmentation of objects. The first stage scans over the image and generates the proposals, and the second stage predicts the class and box offset and produces a binary mask in parallel \cite{b30}. In these two stages, the Mask-RCNN architecture employs three modules: backbone, Region Proposal Network (RPN), and ROI.

The feature maps are constructed in the first stage by extracting image features of various scales using the backbone network such as ResNet (deep residual networks) \cite{b31}, which is also known as the feature extraction network. After that, the obtained feature map is sent to the RPN, which generates proposals. %The RPN has two branches: the first predicts the bounding box, while the second uses SoftMax to do binary classification of foreground and background, with the foreground class implying the presence of a target object in the box. Then non-maximum suppression (NMS) algorithm is used to generate the final proposals (ROIs). Note that the RoIs proposed at this point are obtained from the anchor approach \cite{b32}. 
In the second stage, the corresponding target features of the shared feature maps are extracted by mapping ROIs to feature layers. The RoIAlign is used to modify the feature map to a fixed-size feature map. Finally, the task of mask prediction is completed through FCN branch, and object classification and bounding box regression are completed by two branches of Fully Connected (FC) layers.

% \begin{figure}[htbp]
% \centerline{\includegraphics[width=8.5cm,height=5.5cm]{Figs/Mask.png}}
% \caption{Architecture of Mask-RCNN.}
% \label{fig}
% \end{figure}

\subsubsection{Training strategy}
Deep learning approaches, in general, require a significant amount of data to be trained properly; otherwise, these techniques could fail to yield high accuracy. Therefore, due to the lack of training data and the purpose of applying Mask-RCNN as a deep learning-based approach, we apply transfer learning. We trained all of the layers including the RPN, classifier and mask head of our model network using pre-trained weights from the Common Objects in Context (COCO) \cite{b33} dataset. In addition to transfer learning, we used data augmentation process to address the issue of a limited number of real-world mounds to improve the recognition rate of our model. In this way, a range of some augmentation techniques were used to increase the diversity of the original training dataset.

%horizontal flips, cropping, gaussian blur, adjusting the contrast and illumination to generate brighter and darker images, adding gaussian noise, and affine transformations were used to increase the diversity of the original training dataset.

Once Mask-RCNN is trained through the adaptation of the preceding methodologies, it is fitted to new datasets using a multi-loss function throughout the learning step. As shown in equation (1), the goal is to optimize model parameters by minimizing a multi-tasking loss function that incorporates a three-module combination loss: classification, localization, and segmentation.

\begin{equation} L = L_{cls}+L_{box}+L_{mask},\label{eq}\end{equation}
In this equation, \begin{math}L_{cls}\end{math} represents the loss of classification, \begin{math}L_{box}\end{math} represents the loss of prediction bounding box, and \begin{math}L_{mask}\end{math} represents the loss of mask.

Based on the Mask-RCNN network, the mask branch contains a \begin{math}K m^2\end{math}-dimensional output for each identified ROI that encodes \begin{math}K\end{math} binary masks with a resolution of $m \times m$, representing \begin{math}K\end{math} number of classes \cite{b30}. Thus, \begin{math}L_{mask}\end{math} is defined as the average binary cross-entropy loss on the k-th mask, which is calculated using per-pixel sigmoid on \begin{math}mask_{k}\end{math}, as defined below:

\begin{equation} L_{mask} = Sigmoid(mask_{k})\label{eq}\end{equation}
Local segmentation is done based on annotated patches of planting blocks for the whole terrain. The number of mounds and the ratio of the other three objects in each patch are then used as input to the local count correction.

\subsection{Patch-level correction}
The purpose of this stage is to accurately predict the number of mounds in a given orthomosaic representing a planting block. In our first stage, local image detection and segmentation is used to detect visible mounds. However, the number of visible mounds in an orthomosaic rarely corresponds to the actual number of planted seedlings, due to multiple factors, such as occlusion caused by woody debris or tree from neighboring zone (see Fig.5 (a) and (b)), and destroyed mounds by water flow (see Fig. 5 (c)).

\begin{figure}[htbp]
\centerline{
    \subfloat[]{\label{figur:2}\includegraphics[scale=0.18]
    {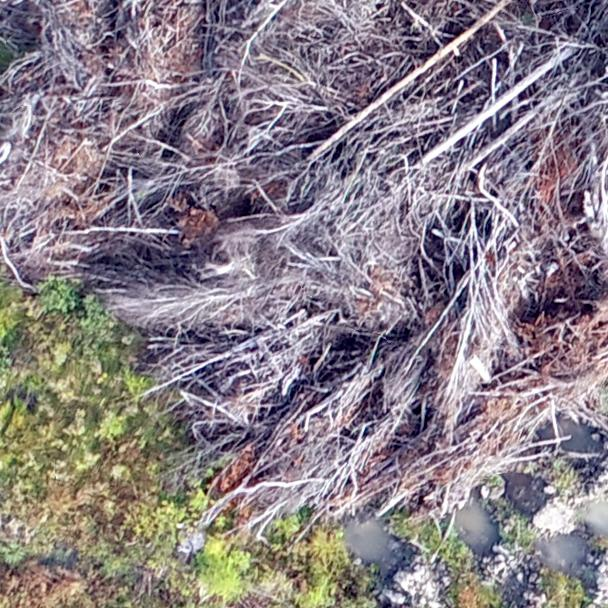}}
    \
    \subfloat[]{\label{figur:2}\includegraphics[scale=0.18]
    {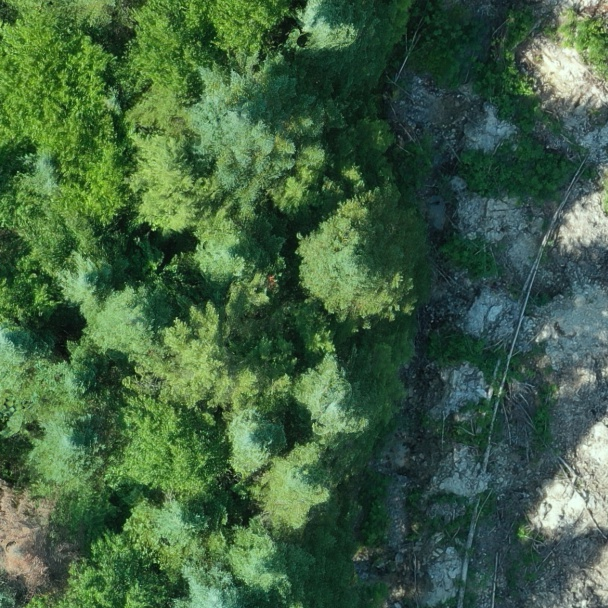}}
     \
    \subfloat[]{\label{figur:2}\includegraphics[scale=0.18]
    {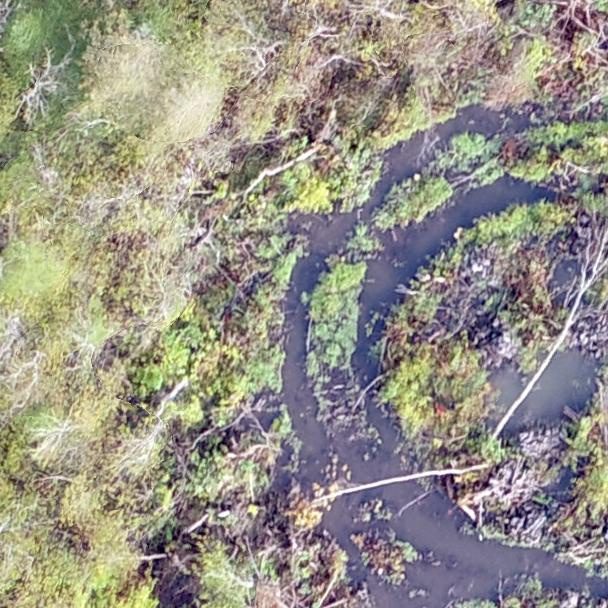}}
    }
\caption{Examples of mound occlusion and destruction. (a) Occlusion due to the
presence of woody debris. (b) Presence of tree from neighboring zone. (c) Destroyed mounds by water flow due to heavy rain.}
\label{fig} 
\end{figure}

Therefore, the number of detected mounds in a local patch is generally underestimated when relying only on detection techniques. To reduce this error, we use regression algorithms based on Mask-RCNN counting results from the previous stage. The objective of regression analysis in this context is to map a function \begin{math}X \rightarrow Y\end{math}, with \begin{math}X\end{math} and \begin{math}Y\end{math} for \begin{math}N\end{math} sample images specified as follows:

\begin{equation}
\begin{aligned}
&\mathrm{X}=\left\{x_{i}^{j}\right\}, \mathrm{j} \in(1,4), \mathrm{i}=1,2, \ldots, \mathrm{N} \\
&\mathrm{Y}=\left\{\mathrm{y}_{\mathrm{i}}\right\}, \mathrm{i}=1,2, \ldots, \mathrm{N}
\end{aligned}
\label{eq}\end{equation}
where \begin{math}x^{1}\end{math} represents the number of detected mounds and \begin{math}x^{2}\end{math}, \begin{math}x^{3}\end{math}, \begin{math}x^{4}\end{math} represent the ratios of trees, water and debris respectively, and \begin{math}Y\end{math} is defined as a corresponding ground-truth.

We investigated regression methods including, linear, Support Vector Regression (SVR), lasso, and Multilayer Perceptron (MLP) as pre-trained models on one orthomosaic in order to find the best predictor according to the best Relative Counting Precision (RCP) and perform it as a regression prediction on patches of any given block.

To determine the final total number of mounds for each block, we use the outcomes of the algorithm predictions for each patch as follow:

\begin{equation}
\begin{aligned}
\text {Count}_{\text {final}}\left(\text {Block}_{i}\right)=\sum_{j=1}^{M}\left(P_{j}\right),
\end{aligned}
\label{eq}\end{equation}
where \begin{math}P_{j}\end{math} represents the j-th patch for a given Block(\begin{math}i\end{math}).

% \begin{figure}[htbp]
% \centerline{
%     \subfloat[]{\label{figur:2}\includegraphics[scale=0.18]
%     {Figs/debries.png}}
%     \
%     \subfloat[]{\label{figur:2}\includegraphics[scale=0.18]
%     {Figs/tree.png}}
%      \
%     \subfloat[]{\label{figur:2}\includegraphics[scale=0.18]
%     {Figs/water flow.png}}
%     }
% \caption{Examples of mound occlusion and destruction. (a) Occlusion due to the
% presence of woody debris. (b) Presence of tree from neighboring zone. (c) Destroyed mounds by water flow due to heavy rain.}
% \label{fig} 
% \end{figure}

\section{Experiments and results}

\subsection{Dataset construction}
To provide input data, a total number of 20 orthomosaics reconstructed from UAV images from different zones with varying characteristics were used. We divided our images into two distinct groups as follows:
\begin{itemize}
\item \textbf{Group 1}: consists of 3  training orthomosaics that were manually annoated for local image segmentation training.

\item \textbf{Group 2}: includes 18 testing orthomosaics used to evaluate the performance for the entire framework. \end{itemize}

The aerial multispectral images were taken with a drone equipped with a high-resolution sensor set vertically and images captured with a high overlap percentage to maximize orthomosaic reconstruction quality at a height of 120 meters. Because the sensor used produced images with a high resolution of 23610 × 18151, we performed a patch-based approach to trim orthomosaic and create non-overlapped patches with regular and stable pixel sizes of 608 × 608. As a result, 1352 patches were employed in total for the model. The data was then processed before being fed into the local instance segmentation method for training. The region of interest was carefully investigated and the ground-truth of mounds and other objects including tree, water and woody debris was manually annotated using the open source VGG Image Annotator (VIA) tool \cite{b34}. An example of patch cropped to a fixed cell size of 608 × 608 pixels, showing manually annotated objects is presented in Fig. 6.

%\begin{figure}[htbp]
%\centering
%     \subfloat[\label{1a}]{%
%     \includegraphics[width=2.5cm,height=4cm]{Figs/Zone1.png}}

% \centerline{
% %    \subfloat[]{\label{figur:1}\includegraphics[width=2.5cm,height=4cm]
% %    {Zone1.png}}
%     \
%   \subfloat[]{\label{figur:2}\includegraphics[scale=0.25]
%     {Figs/Patch1.png}}
%   \
%     \subfloat[]{\label{figur:2}\includegraphics[scale=0.249]
%     {Figs/Bbox.png}}
%     }
% \caption{(a) Example of one orthomosaic and a sample patch cropped to a fixed dimension. (b) manually annotated objects (mound, tree, water, debris).}
% \label{fig} 
%\end{figure}

\begin{figure}[htbp]
\centerline{
    \subfloat[]{\label{figur:2}\includegraphics[scale=0.4]
    {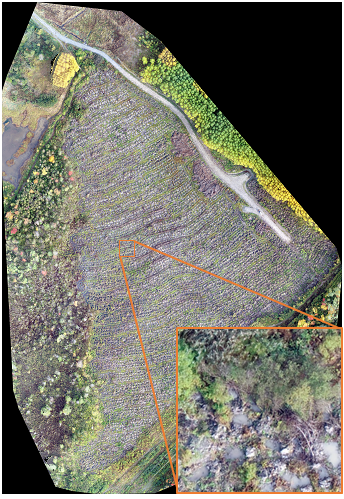}}
    \
    \subfloat[]{\label{figur:2}\includegraphics[scale=0.326]
    {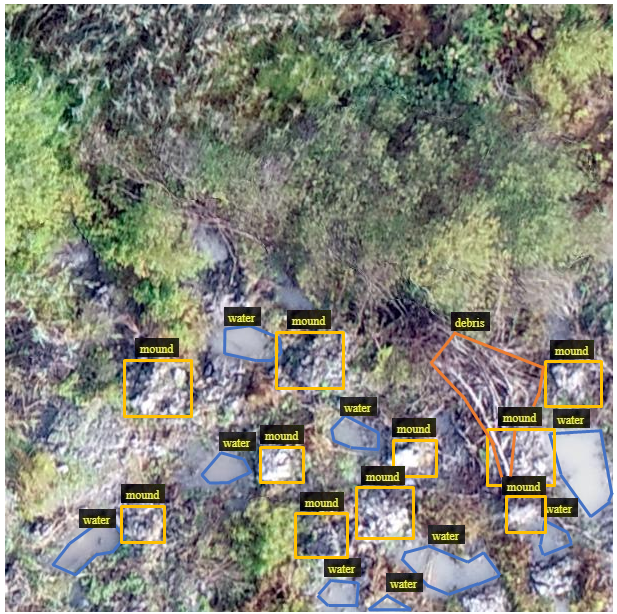}}
    }
\caption{(a) Example of one orthomosaic and a sample patch cropped to a fixed dimension. (b) Manually annotated objects (mound, tree, water, debris).}
\label{fig} 
\end{figure}

\subsection{Evaluation metric}
% The performance of the visual object instance detection and segmentation experiment is evaluated through mean Average Precision (mAP), which is calculated using the confusion matrix and the following sub metrics of precision (P) and recall (R):

% \begin{equation}
% \begin{aligned}
% &P =\frac{TP}{TP+FP} \\
% \end{aligned}
% \label{eq}\end{equation}

% \begin{equation}
% \begin{aligned}
% &R =\frac{TP}{TP+FN}
% \end{aligned}
% \label{eq}\end{equation}
% where \begin{math}TP\end{math} indicates the number of True Positives, \begin{math}FP\end{math} indicates the number of False Positives, and \begin{math}FN\end{math} indicates the number of False Negatives.

% The area under the Precision-Recall curve is represented by Average Precision (AP) that is calculated for each class using the Intersection over Union (IoU) as a metric for measuring the overlap between ground-truth and the predicted mask as below: 

% \begin{equation}
% \begin{aligned}
% AP=\sum_{i=0}^{n-1}\left(R_{i+1}-R_{i}\right) P_{\text {interp }}\left(R_{i+1}\right),
% \end{aligned}
% \label{eq}\end{equation}
% where \begin{math}Pinterp(R)\end{math} is the precision interpolated at a certain recall level.

% Given that \begin{math}AP_i^x\end{math} is the average precision for a given IOU threshold of \begin{math}x\end{math} and class \begin{math}i\end{math}, mAP for \begin{math}N\end{math} number of classes is defined by:

% \begin{equation}
% \operatorname{mAP}=\frac{\sum_{i=1}^{N} AP_{i}^{x}}{N}
% \label{eq}\end{equation}

%Finally, 
We used the relative counting precision metric to measure the overall system performance by evaluating the regression predictors and obtaining a final counting estimate, as shown below:

\begin{equation}
RCP= 1 - \left| \dfrac{\#predicted \ mound- \#gt}{\#gt}\right|
\label{eq}\end{equation}
where  \emph{\#predicted-mound} represents the predicted number of mounds and \begin{math}\#gt\end{math} represents the number of mounds from the ground-truth.

\subsection{Experimental results}
The process of counting mounds on a new planting block comprised two steps:

\begin{table*}[htbp]{\centering}
\caption{Quantitative results of our proposed approach. \\  GroundTruth refers to the final number of plant seedlings planted in a block, local segmentation-based count is the number of mounds detected and segmented using local image segmentation method, count is the number of locally corrected mounds and RCP corresponds to the relative counting precision. Average precision measure represents the average over all precision values, but the overall result indicates the counting precision when the entire number of mounds in the dataset is taken into account.}
\begin{center}
\begin{tabular}{|c|c|c|c|c|c|c|c|c|c|c|c|}
\hline
\textbf{}&\textbf{}&\multicolumn{2}{|c|}{\textbf{Local}}&\multicolumn{8}{|c|}{\textbf{Patch-level corrected count}} \\
\cline{5-12} 
\textbf{Orthomosaic} &\textbf{GroundTruth} &\multicolumn{2}{c|}{\textbf{\textit{segmentation-based count}}} & \multicolumn{2}{c|}{\textbf{\textit{Linear}}}& \multicolumn{2}{c|}{\textbf{\textit{SVR}}}&  \multicolumn{2}{c|}{\textbf{\textit{Lasso}}}& \multicolumn{2}{c|}{\textbf{\textit{MLP}}} \\
\cline{3-12}
\textbf{} &\textbf{} &\textbf{\textit{Count}}& \textbf{\textit{RCP}} & \textbf{\textit{Count}}& \textbf{\textit{RCP}}& \textbf{\textit{Count}} &\textbf{\textit{RCP}} & \textbf{\textit{Count}}& \textbf{\textit{RCP}}& \textbf{\textit{Count}}& \textbf{\textit{RCP}} \\
\hline
Block 01 & 16450 & 14458 & 88\% & 15820 & 96\% & 15180 & \emph{92\%} & 15504 & 94\% & 15828 & 96\% \\
\cline{1-12}
Block 02 & 2650 & 2609 & 98\% & 2816 & 94\% & 2760 & \textbf{96\%}$^{\mathrm{a}}$ & 2851 & 92\% & 2780 & 95\% \\
\cline{1-12}
Block 03 & 750 & 712 & 95\% & 724 & 97\% & 737 & \textbf{98\%} & 771 & 97\% & 728 & 97\% \\
\cline{1-12}
Block 04 & 800 & 784 & 98\% & 851 & 94\% & 844 & \emph{94\%} & 891 & 89\% & 837 & 95\% \\
\cline{1-12}
Block 05 & 2350 & 2233 & 95\% & 2495 & 94\% & 2436 & \textbf{96\%} & 2562 & 91\% & 2462 & 95\% \\
\cline{1-12}
Block 06 & 1700 & 1513 & 89\% & 1623 & 95\% & 1600 & \emph{94\%} & 1683 & 99\% & 1621 & 95\% \\
\cline{1-12}
Block 07 & 2050 & 1853 & 90\% & 1868 & 91\% & 1879 & \emph{92\%} & 1933 & 94\% & 1864 & 91\% \\
\cline{1-12}
Block 08 & 3950 & 3443 & 87\% & 3676 & 93\% & 3571 & \emph{90\%} & 3649 & 92\% & 3629 & 92\% \\
\cline{1-12}
Block 09 & 6847 & 6632 & 97\% & 7041 & 97\% & 6915 & \textbf{99\%} & 7091 & 96\% & 6923 & 99\%  \\
\cline{1-12}
Block 10 & 30200 & 28301 & 94\% & 28973 & 96\% & 29145 & \textbf{97\%} & 30107 & 99.7\% & 28733 & 95\% \\
\cline{1-12}
Block 11 & 2950 & 2742 & 93\% & 2778 & 94\% & 2797 & \emph{95\%} & 2894 & 98\% & 2765 & 94\% \\
\cline{1-12}
Block 12 & 25450 & 24251 & 95\% & 2765 & 96\% & 25447 & \textbf{99.99\%} & 25994 & 98\% & 25848 & 98\%  \\
\cline{1-12}
Block 13 & 7400 & 6658 & 90\% & 7825 & 94\% & 7551 & \textbf{98\%} & 8079 & 91\% & 7824 & 94\% \\
\cline{1-12}
Block 14 & 5250 & 5009 & 95\% & 5620 & 93\% & 5468 & \textbf{96\%} & 5751 & 90\% & 5563 & 94\%  \\
\cline{1-12}
Block 15 & 3557 & 3424 & 96\% & 3636 & 98\% & 3653 & \textbf{97\%} & 3842 & 92\% & 3643 & 98\% \\
\cline{1-12}
Block 16 & 5150 & 4320 & 84\% & 5362 & 96\% & 5032 & \textbf{98\%} & 5418 & 95\% & 5331 & 96\% \\
\cline{1-12}
Block 17 & 4900 & 4759 & 97\% & 5164 & 95\% & 5025 & \textbf{97\%} & 5236 & 93\% & 5128 & 95\% \\
\cline{1-12}
Block 18 & 2650 & 2267 & 86\% & 2478 & 94\% & 2492 & \emph{94\%} & 2670 & 99.2\% & 2471 & 93\% \\
\cline{1-12}
Overall result & 125054 & 115968 & 93\% & 101515 & 81\% & 122532 & \textbf{98\%} & 126926 & 99\% & 123978 & 99\% \\
\cline{1-12}
Average precision & \multicolumn{2}{c|}{\textbf{\textit{}}} & 93\% & ~ & 95\% & ~ & \textbf{96\%} & ~ & 94\% & ~ & 95\% \\
\cline{1-12}
\multicolumn{4}{l}{$^{\mathrm{a}}$Highlighted numbers contribute more significantly to overall precision.}
\end{tabular}
\label{tab1}
\end{center}
\end{table*}

\begin{enumerate}
\item Applying the local image segmentation model to detect visible mounds and quantify the presence of trees, debris and water.

\item Performing patch-level correction using a regression function. \end{enumerate}

Note that we only trained our local image segmentation using three annotated orthomosaics from Group 1. To ensure that the models work properly and verify the performance of our proposed method, we used 18 orthomosaics from Group 2 that had not been utilized in the training processes.

%Table I shows the quantitative result on test data for constructing a local image segmentation model. We included mAP at 50\% (overlap threshold to defined \begin{math}TP\end{math} bounding boxes) and 75\% over all test patches. It can be observed from table I that the quantitative result of the mAP with the IoU threshold of 0.5 was 52\%. Therefore we used this threshold to produce the results of our local image segmentation on test dataset.

Fig. 7 depicts one sample patch and its corresponding qualitative result. The number of mounds and the ratio of other detected and segmented objects were then used as input features in the second step of pre-trained regression algorithm to produce final mound counting. 

\begin{figure}[htbp]
\centerline{
%    \subfloat[]{\label{figur:1}\includegraphics[width=2.5cm,height=4cm]
%    {Zone1.png}}
    \
    \subfloat[]{\label{figur:2}\includegraphics[scale=0.55]
    {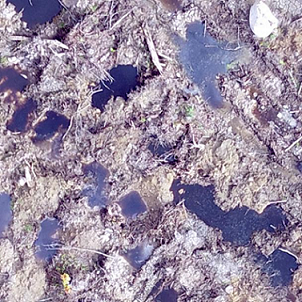}}
    \
    \subfloat[]{\label{figur:2}\includegraphics[scale=0.196]
    {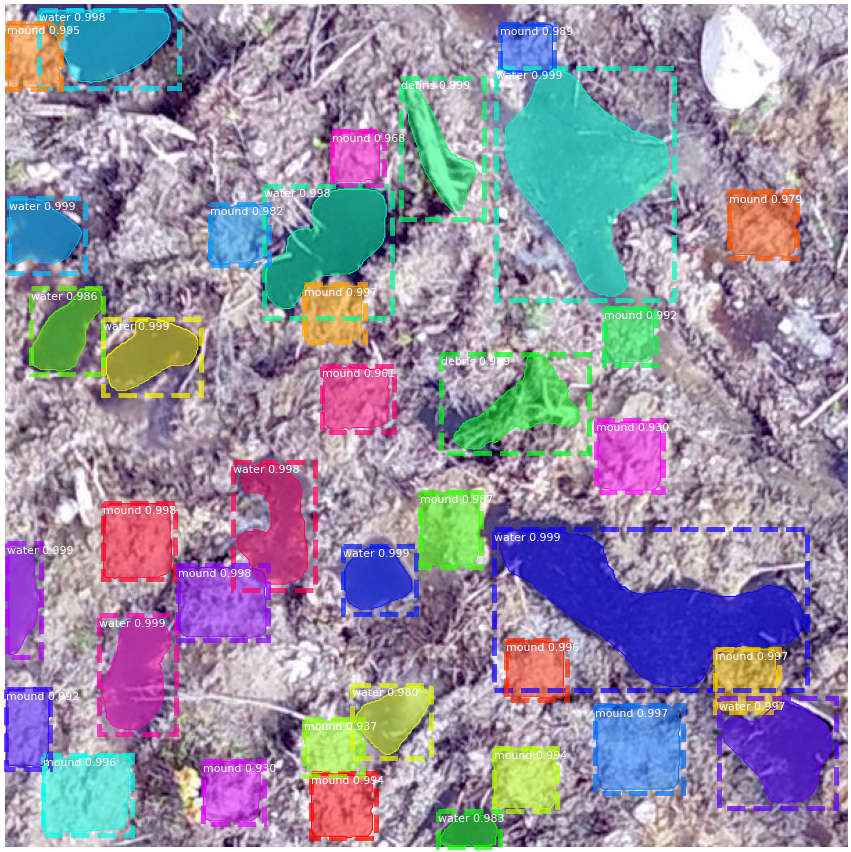}}
    }
\caption{(a) Example of one patch. (b) Corresponding qualitative result.}
\label{fig} 
\end{figure}

% \begin{table}[htbp]
% \caption{Quantitative result of mAP}
% \begin{center}
% \begin{tabular}{|c|c|c|}
% \hline
% \textbf{Number of Patches}&\textbf{mAP 50}&\textbf{mAP 75} \\
% \cline{1-3} 
% \hline
% 61 & 52\% & 16\% \\
% \cline{1-3}
% \end{tabular}
% \label{tab1}
% \end{center}
% \end{table}

Table I shows the quantitative results of our proposed approach. According to the findings, the RCP of local image segmentation method is 93\%, which indicates that our instance segmentation method has the ability to detect and segment the planting microsites efficiently. However, we could significantly improve the average detection precision of counting mounds by applying patch-level correction methods. From table I, this improvement reached 96\% by performing SVR. Compared with other regression methods, SVR yields the greatest results since it employs a kernel function to concurrently minimize prediction errors and model complexity. 

Although the experimental results show the efficiency of our strategy, this study is subject to several challenges. According to the results of local-segmentation mound counting, the RCP for blocks 01, 06, 08, 16, and 18 are less than 90\% compared to the other ones, which are equal to or greater than 90\%. This can be caused by the fact that the new plantation block may exhibit many unseen properties when we test our object-pixel-wise instance detector. This could include mound shapes and block characteristics that the detector was not exposed to during training. In addition, table I indicates that while our framework was able to improve the final results by applying SVR, the outcomes for some blocks are lower than others. As can be seen from block 08, our correction method could only correct 128 more mounds than the local approach and bringing the total number of mounds from 3443 up to 3571. The reason of underestimating the number of corrected mounds in comparison to the ground-truth derived from the challenging situations on the captured images of the related blocks. For instance, because the images of block 08 were taken under dry conditions, the texture of mounds is similar to surrounding regions, which makes object detection very challenging.

Despite aforementioned constraints, the results demonstrate that using the two stages consecutively can result in an average improvement of 3\%. The comparison findings show that applying the patch-level correction module always enables an improvement in RCP compared to segmentation-based count. In addition, the RCP of the manual method is around 85\% which confirms that our framework outperforms the traditional method of counting mounds.

\section{Conclusion}
In this paper, we proposed a new computer vision framework to detect and segment multiple objects, followed by counting mechanically created mounds for planting from UAV images. The proposed system consists of a hybrid approach, which combines a local image segmentation method with patch-level count correction. In this regard, the objective of local image segmentation method was to segment pixel-level visual mounds. The accurate number of final mounds was then determined using the patch-level correction approach. The experimental results demonstrate that our approach is effective in dealing with a variety of difficult conditions involving environmental factors and the existence of multiple objects in each terrain patch. That is, the local segmentation approach leverages visual mounds from aerial images to detect a preliminary count, which is subsequently improved by the patch-level correction method.
According to qualitative and quantitative performance assessments, our approach outperforms traditional counting methods (i.e. field work) in terms of precision, and reduces the financial cost of planning planting operations significantly.

%\section*{References}

\vspace{12pt}
\end{document}